\documentclass{article}
\usepackage{spconf,amsmath,graphicx}

\usepackage{amssymb}
\usepackage{bm}
\usepackage{booktabs}
\usepackage{cite}
\usepackage{url}

\title{G-MASt3R-SfM: Graph-based View Pruning and Multi-stage Optimization for Robust SfM}
\name{Toshiki Watanabe, Shintaro Ito, Natsuki Takama, Koichi Ito, and Takafumi Aoki}
\address{Graduate School of Information Sciences, Tohoku University, Japan}

\begin{document}
\ninept
\maketitle
\begin{abstract}
  Structure from Motion (SfM) is essential for multi-view 3D reconstruction, however, its accuracy heavily relies on the accuracy of image matching.
  While the recent correspondence matching method, MASt3R, enables robust matching even under challenging conditions, it tends to generate incorrect correspondences for non-overlapping image pairs.
  Consequently, existing SfM methods using MASt3R, such as MASt3R-SfM, suffer from significant degradation in pose estimation accuracy as they incorporate these unreliable matches directly into optimization.
  To address this issue, we propose G-MASt3R-SfM, a novel SfM pipeline that enhances robustness through two key modules.
  First, the Graph-based View Pruning (GVP) module constructs a scene graph from matching confidence and geometrically prunes outlier views.
  Second, the Multi-Stage Optimization (MSO) module progressively refines camera parameters by expanding the optimization scope from local consistency to the global consistency.
  Experiments on the ETH3D dataset demonstrate that our method achieves state-of-the-art accuracy in both camera pose estimation and 3D reconstruction, effectively suppressing noise caused by outliers.
\end{abstract}
\section{Introduction}
\label{sec:intro}

Multi-view 3D reconstruction, which recovers the 3D shape of a target from multiple images captured by cameras, has been widely adopted in applications such as digital archiving of cultural heritage and 3D map creation for autonomous driving due to its simplicity and low cost.
To recover 3D shapes with high accuracy, it is essential to accurately estimate camera parameters, such as camera pose and focal length, from the multiple images.
Structure from Motion (SfM) \cite{CV} serves as the standard approach for this purpose.
Recent advancements in deep learning have led to significant improvements in the accuracy of multi-view 3D reconstruction methods \cite{multiview}.
However, most of these methods assume that camera parameters are known; consequently, their reconstruction accuracy relies heavily on the estimation accuracy of camera parameters by SfM.

COLMAP \cite{johannes2016colmap}, the de facto standard SfM tool, employs detector-based image matching \cite{arandjelovic2012rootsift}.
While localization accuracy for feature points is high, these methods often struggle to find sufficient correspondences in texture-less regions or across images with significant illumination changes.
In contrast, deep learning-based methods such as MASt3R \cite{vincent2024mast3r} enable dense and robust matching even with limited overlap, leading to applications in SfM like MASt3R-SfM \cite{bardienus2024mast3r-sfm}.
However, applying MASt3R to SfM presents several challenges.
Although MASt3R is powerful in image correspondence between two images, it tends to output correspondences even for non-overlapping image pairs.
Since MASt3R-SfM incorporates such unreliable correspondence results into camera parameter estimation without filtering, it often fails to achieve global consistency during bundle adjustment, resulting in significant accuracy degradation.
Furthermore, the dense point clouds output by MASt3R can have lower localization accuracy than feature-based methods.
Thus, as the number of views increases, maintaining geometric consistency between views becomes difficult, further reducing estimation accuracy.
To achieve accurate SfM while leveraging MASt3R's strengths, it is essential to appropriately filter view connectivity for camera parameter estimation and employ an optimization strategy that ensures multi-view consistency.

In this paper, we propose G-MASt3R-SfM, a novel camera parameter estimation method utilizing a scene graph and multi-stage optimization to enhance the accuracy and robustness of MASt3R-based SfM.
Our method consists of two main modules.
First, the Graph-based View Pruning (GVP) module constructs a scene graph where nodes represent views based on MASt3R's confidence scores, and prunes outlier views and erroneous connection groups by graph structure analysis.
This module effectively eliminates inappropriate views from the estimation process.
Second, the Multi-Stage Optimization (MSO) module refines camera parameters by performing bundle adjustment on the selected views, progressively expanding the optimization scope from local consistency to global consistency.
Experiments on the ETH3D dataset \cite{schops2017eth3d} demonstrate that our method achieves higher accuracy and stability in camera parameter estimation across diverse scenes compared to existing methods.

\begin{figure*}[t]
  \centering
  \includegraphics[width=\linewidth]{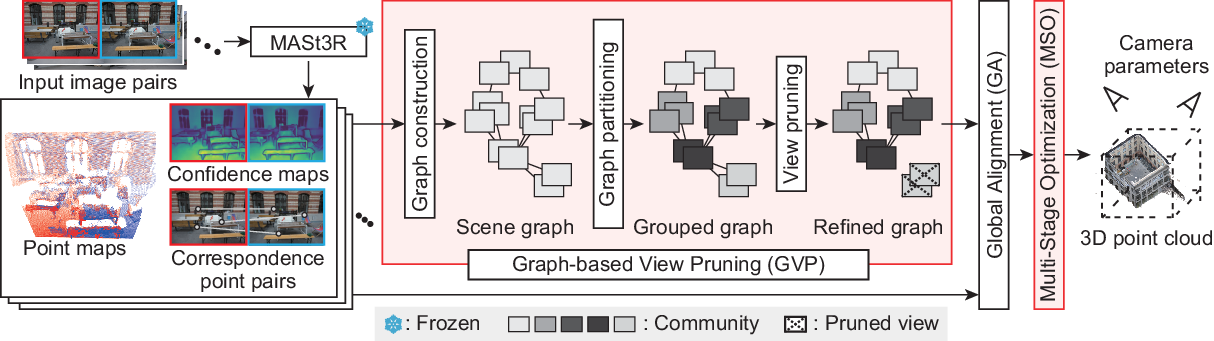}
  \caption{The overall pipeline of the proposed G-MASt3R-SfM.
  The modules highlighted in red (GVP and MSO) represent our novel contributions to the standard MASt3R-SfM pipeline.
  Utilizing correspondences and confidence maps from MASt3R, the GVP module constructs a scene graph to filter out unreliable views, followed by the MSO module which refines camera parameters.}
  \label{fig:proposed}
\end{figure*}

\section{Related Work}
\label{sec:related_work}

In this section, we give a brief overview of camera parameter estimation from images, focusing on SfM and rapidly evolving 3D foundation models.

\noindent
{\bf Structure from Motion (SfM)} ---
SfM is a technique that simultaneously estimates 3D structure and camera parameters using correspondences between images.
COLMAP \cite{johannes2016colmap}, the standard SfM pipeline, relies on detector-based matching \cite{detone2018sp,Aliked} such as SIFT \cite{lowe2004sift}.
However, it faces challenges in texture-less regions or scenes with significant illumination changes where sufficient correspondences cannot be obtained, leading to degraded estimation accuracy.
In contrast, DFSfM \cite{DFSfM} improves robustness by incorporating detector-free dense matching methods \cite{LoFTR} into SfM.
However, dense geometric information is lost during the process of aggregating correspondences for computational efficiency.
VGGSfM \cite{VGGSfM} employs a tracker that simultaneously estimates correspondences across all views; however, tracking becomes difficult between images with limited overlap, thereby restricting applicable scenes.

\noindent
{\bf 3D Foundation Models} ---
Recently, foundation models trained on large-scale 3D datasets have been applied to various tasks, including camera parameter estimation.
DUSt3R \cite{wang2024dust3r} estimates relative camera poses without explicit matching by directly regressing point maps from image pairs.
MASt3R \cite{vincent2024mast3r} extends DUSt3R to achieve more accurate image matching and geometric estimation.
However, since these methods are fundamentally based on pairwise inference, post-processing such as global optimization is indispensable to guarantee multi-view consistency.
To address this issue, MUSt3R \cite{MUSt3R} and Spann3R \cite{Spann3R} attempt to extend these models to multi-view settings by processing input images sequentially.
Similarly, VGGT \cite{VGGT} learns multi-view geometric relationships using Transformers.
However, these methods prioritize computational efficiency and often lack the rigorous geometric verification inherent to standard SfM pipelines.
Consequently, as the number of views increases, estimation errors accumulate, leading to a loss of global consistency.

\section{G-MASt3R-SfM}
\label{sec:method}

As mentioned above, to leverage the high matching capability of MASt3R \cite{vincent2024mast3r} while preventing SfM failure caused by incorrect correspondences and outliers, it is effective to organize view connectivity and perform stepwise optimization.
In this paper, we introduce the GVP module and the MSO module to MASt3R-SfM \cite{bardienus2024mast3r-sfm}.
Fig. \ref{fig:proposed} illustrates the overall pipeline of our proposed G-MASt3R-SfM.
First, we construct a scene graph based on MASt3R outputs and prune geometrically inconsistent groups of views by graph structure analysis.
Next, for the selected views, we refine camera parameters while maintaining global consistency by progressively expanding the optimization scope based on the community structure of the graph.
In the following, we describe the overview of MASt3R-SfM and the details of two proposed modules.

\subsection{MASt3R-SfM}

MASt3R-SfM \cite{bardienus2024mast3r-sfm} integrates dense point clouds and correspondence information output by MASt3R \cite{vincent2024mast3r} to perform SfM based on global optimization.
The process primarily consists of two stages: Global alignment and bundle adjustment.
First, the Global Alignment (GA) module aligns the point maps of each view to a common world coordinate system.
For all view pairs $(n,m) \in \bm{E}$, the scale and extrinsic parameters of each view are optimized to minimize the 3D position error $e_{\rm{pos}}$ of the set of correspondence pairs $\bm{M}_{n,m}$.
Let $c^{\bm{p}}$ be the confidence of correspondence pair $\bm{p} \in \bm{M}_{n,m}$, and $\bm{X}_n^{\bm{p}}$ be the 3D coordinate of $\bm{p}$ in view $n$.
The error $e_{\rm{pos}}$ is defined by
\begin{equation}
  e_{\rm{pos}} = \sum_{(n,m) \in \bm{E}}\sum_{\bm{p} \in \bm{M}_{n,m}} c^{\bm{p}}
  \bigl( \| \bm{X}^{\bm{p}}_n - \bm{X}^{\bm{p}}_m\| \bigr).
  \label{equ:e_pos}
\end{equation}
Next, all parameters are optimized by bundle adjustment.
Using the results of the GA module as initialization, the intrinsic and extrinsic parameters, as well as depth maps for each view, are optimized to minimize the reprojection error $e_{\rm{rep}}$ across all view pairs.
Let $\pi_n$ be the projection function mapping a 3D point in the world coordinate system to the image plane of view $n$, and $\bm{x}^{\bm{p}}_n$ be the 2D coordinate of the corresponding point $\bm{p}$ in view $n$.
The error $e_{\rm{rep}}$ is given by
\begin{equation}
  e_{\rm{rep}} = \sum_{(n,m) \in \bm{E}}\sum_{\bm{p} \in \bm{M}_{n,m}} c^{\bm{p}}
  \bigl( \| \bm{x}^{\bm{p}}_n - \pi_n(\bm{X}^{\bm{p}}_m) \|
  + \| \bm{x}^{\bm{p}}_m - \pi_m(\bm{X}^{\bm{p}}_n)\| \bigr).
  \label{equ:e_rep}
\end{equation}

\subsection{Graph-based View Pruning (GVP) Module}

Since MASt3R \cite{vincent2024mast3r} may output correspondences even for non-overlapping image pairs, it can degrade SfM accuracy.
The GVP module constructs a scene graph based on geometric verification and removes outlier views through graph structure analysis.
We construct a scene graph where nodes represent images and edges represent the connection strength between them.
Rather than using MASt3R outputs directly, we filter for geometrically consistent correspondences to calculate edge weights.
Specifically, for each pair $(n,m)$, we estimate the relative pose using the focal length derived from MASt3R's point maps and the fundamental matrix computed by RANSAC \cite{RANSAC}.
Using this relative pose, we reproject points and retain only those falling within the image plane as inliers.
Finally, a reliable graph $G$ is constructed by adding edges only for pairs where the sum of inlier confidences $c^{\bm{p}}$ exceeds a threshold, which we set to 1,000 in the experiments.
We apply the Louvain method \cite{Louvain} to the constructed scene graph $G$ to partition nodes into densely connected communities $\{C_1, C_2, \dots\}$.
We assume that outlier views deviating from the scene form small communities that are either isolated or weakly connected to the main component.
To identify them, we embed the graph into a 2D plane using the Spring Layout \cite{spring_layout}, a force-directed algorithm, and evaluate the separation of each community.
Let $dist(u,v)$ be the distance between nodes $u$ and $v$ in the 2D plane, and $Scale$ be the diagonal length of the entire node distribution area.
The separation score $s_i$ for community $C_i$ is defined by
\begin{equation}
  s_i = \frac{\min_{u \in C_i, v \notin C_i} dist(u, v)}{Scale} \cdot \frac{1}{\log(1 + |C_i|)},
\label{equ:score}
\end{equation}
where $|C_i|$ denotes the size of the community.
This score $s_i$ increases as the community becomes more distant from others and smaller in size.
In this paper, communities with $s_i > 1.5$ are considered outliers and removed from the graph.

\subsection{Multi-Stage Optimization (MSO) Module}

\begin{figure}[t]
  \centering
  \includegraphics[width=\linewidth]{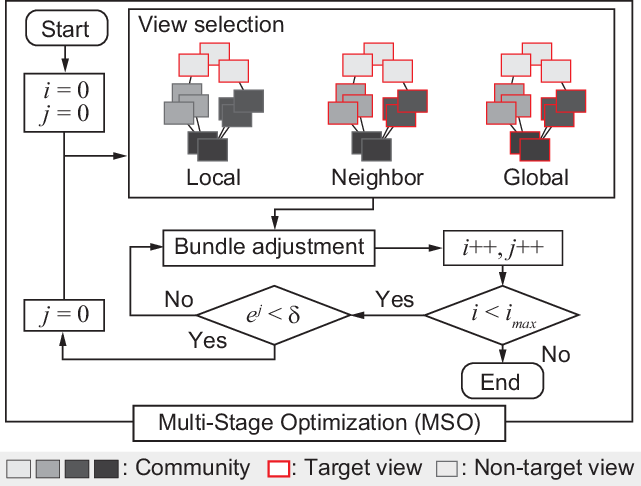}
  \caption{Processing flow of the MSO module.
  The optimization scope acts as a Target view (red outline) and progressively expands from local to global stages.}
  \label{fig:method_module}
\end{figure}

As shown in Fig. \ref{fig:method_module}, after initialization by the GA module on the selected views, we perform iterative bundle adjustment in three stages ($Local$, $Neighbor$, and $Global$) utilizing the graph's community structure.
Instead of optimizing all views simultaneously from the start, we stabilize solution convergence by gradually expanding the optimization scope from local to global consistency.
Let $i$ be the current optimization iteration and $j$ be the repetition count within each stage.
The optimization at each stage is as follows:

\noindent
{\bf Local}:
Optimization is performed independently for each community $C_i$ using only internal views.
This establishes local geometric consistency for each densely connected group of views.

\noindent
{\bf Neighbor}:
To enhance consistency between communities, we optimize each community $C_i$ together with views included in its adjacent communities $Adj(C_i)$, where $Adj(C_i)$ refers to all communities containing nodes connected by edges to nodes within $C_i$.
To reduce computational cost, if the number of target views exceeds half the total number of views $N$, we select $\operatorname{round}(N / 2)$ views via uniform sampling.

\noindent
{\bf Global}:
Bundle adjustment is performed using all views.

Optimization in each stage is performed by minimizing the reprojection error $e_{\rm{rep}}$ in Eq. (\ref{equ:e_rep}).
For convergence criteria, we monitor the ratio $e^j$ between the average error of the last $k$ iterations ($k=5$) and the preceding $l$ iterations ($l=10$), repeating until $e^j < \delta$ ($\delta=0.01$).
Upon satisfying the convergence condition in the $Global$ stage, the process returns to the $Local$ stage and repeats.
The process terminates when the total optimization count $i$ reaches the upper limit $i_{max}$.

\section{Experiments and Discussion}

In this section, we present experiments to demonstrate the effectiveness of our proposed G-MASt3R-SfM.

\subsection{Experimental Settings}

The experimental conditions are summarized below.

\noindent
{\bf Dataset} ---
We use the ETH3D dataset \cite{schops2017eth3d}, which consists of multi-view images and camera parameters captured in 25 indoor and outdoor scenes.
ETH3D provides highly accurate camera parameters as ground truth, obtained by refining COLMAP \cite{johannes2016colmap} estimation results using point clouds measured by a laser scanner and images.
In the experiment, we use 13 scenes out of the 25 scenes.
The number of images varies per scene, ranging from a minimum of 14 to a maximum of 76.
For scenes containing more than 50 images, we subsample half of the images at regular intervals to reduce computational costs.

\noindent
{\bf Baselines} ---
As baselines, we use COLMAP, DFSfM \cite{DFSfM}, VGGSfM \cite{VGGSfM}, VGGT \cite{VGGT}, and MASt3R-SfM \cite{bardienus2024mast3r-sfm}.
For image matching, COLMAP uses RootSIFT \cite{arandjelovic2012rootsift}, and DFSfM uses LoFTR \cite{LoFTR}.
In MASt3R-SfM, the number of optimization iterations for the GA module and bundle adjustment are both set to 500.
For fair comparison, we estimate correspondences for all pairs of input images in all methods.

\noindent
{\bf Implementation Details} ---
We use an NVIDIA GeForce RTX 4090 GPU (24GB) for the experiments.
G-MASt3R-SfM is implemented in PyTorch, based on the official implementation of MASt3R-SfM.
We use Adam \cite{KingmaB14} for optimization and set the total number of optimization iterations  $i_{max}$ to 500.
In the GVP module, the resolution for the Louvain method \cite{Louvain} is set to 1.0, and the repulsion coefficient for the Spring Layout \cite{spring_layout} is set to 0.8.

\noindent
{\bf Evaluation Metrics} ---
The accuracy of camera parameter estimation is evaluated using the Relative Rotation Error (RRE) and Relative Translation Error (RTE), which represent the errors of the estimated rotation matrices and translation vectors, respectively.
As a comprehensive metric, we use the Area Under the Curve (AUC@$\delta$) of the cumulative error curve representing the proportion of pairs where both RRE and RTE are below a threshold $\delta$.
In this experiment, we set $\delta = 5^\circ$.
Additionally, we define the ``SfM rate'' as the ratio of images successfully registered by SfM among all input images, serving as a metric for robustness.
The accuracy of 3D point cloud reconstruction is evaluated using the standard metrics of the ETH3D benchmark: Accuracy (Acc.), Completeness (Cpl.), and F1 score.
``Acc.'' indicates the precision of the reconstructed points based on the distance from the reconstructed point cloud to the ground truth.
``Cpl.'' indicates the coverage of the target shape based on the distance from the ground truth to the reconstructed point cloud.
``F1 score'' is the harmonic mean of ``Acc.'' and ``Cpl.'', representing the overall quality of the reconstruction.

\subsection{Ablation Study}

\begin{figure}[t]
  \centering
  \includegraphics[width=\linewidth]{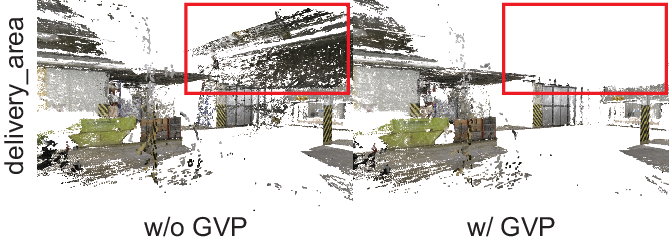}
  \caption{Effect of the GVP module on 3D reconstruction quality.
  The proposed method eliminates artifacts observed in the baseline (left).}
  \label{fig:result_ablation}
\end{figure}

In this experiment, we verify the effectiveness of the GVP and MSO modules that are key modules of our proposed method.
For comparison, we establish a baseline (denoted as ``Random'') that selects views with the highest sum of correspondence confidence relative to a randomly chosen anchor view.
Note that the number of views in the $Local$ and $Neighbor$ stages for ``Random'' is set to be equivalent to that of the proposed method.
Table \ref{tab:results_ablation} summarizes the experimental results.
First, we analyze the impact of including the GVP module.
Regardless of the view selection strategy, introducing GVP leads to drastic improvements in both RRE and RTE.
This indicates that proactively pruning geometrically inconsistent views is effective for improving SfM accuracy.
Next, we compare the view selection strategies.
When GVP is not applied, no clear performance difference is observed between ``Random'' and the proposed method.
On the other hand, when combined with GVP, the proposed method achieves higher accuracy than ``Random'', yielding the best results across all metrics.
This result demonstrates that GVP constructs a highly reliable graph, and the stepwise optimization based on its community structure is effective.
Fig. \ref{fig:result_ablation} shows the impact of GVP on the quality of 3D reconstruction.
Without GVP, point clouds are generated in wrong positions due to incorrect camera parameters.
In contrast, applying GVP removes these artifacts, resulting in an accurate 3D reconstruction of the target object.

\begin{table}[t]
  \centering
  \caption{Ablation study on view selection and pruning.}
  \label{tab:results_ablation}
  \begin{tabular}{cccccc}
    \toprule
    \multicolumn{2}{c}{View selection} & GVP & RRE $\downarrow$ & RTE $\downarrow$ & SfM rate \\
    \cmidrule(lr){1-2}
    Random     & Ours                  & module & [deg.]           & [deg.]           & [\%]     \\
    \midrule
    \checkmark &                       &         & 2.173            & 2.307            & 100      \\
    \checkmark &                       & \checkmark & 0.479            & 0.992            & 97       \\
               & \checkmark            &         & 2.231            & 2.247            & 100      \\
               & \checkmark            & \checkmark & \textbf{0.474}   & \textbf{0.978}   & 97       \\
    \bottomrule
  \end{tabular}
\end{table}

\subsection{Scene Graph}

\begin{figure}[t]
  \centering
  \includegraphics[width=\linewidth]{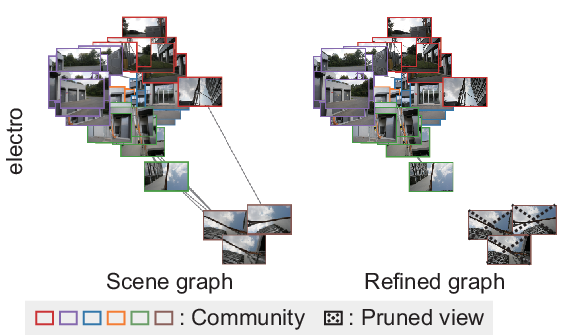}
  \caption{Examples of scene graphs constructed by our GVP module.}
  \label{fig:graph}
\end{figure}

Fig. \ref{fig:graph} shows an example of a scene graph constructed by the GVP module.
In this scene, a single community has been removed based on community detection and geometric evaluation.
Upon inspecting the images within the removed community, we observe that they are dominated by sky regions and exhibit minimal overlap with the main group of images.
This result demonstrates that the GVP module effectively detects and eliminates inappropriate views that would otherwise lead to a degradation in SfM accuracy.

\begin{figure}[t]
  \centering
  \includegraphics[width=\linewidth]{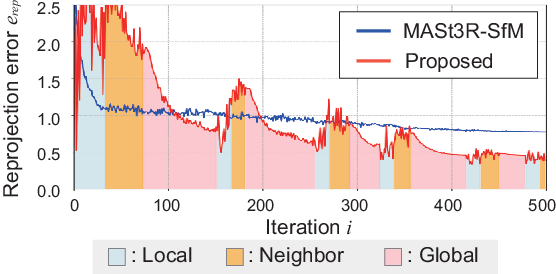}
  \caption{Comparison of reprojection error convergence.
  The proposed method achieves a lower final error compared to MASt3R-SfM through the Multi-Stage Optimization (MSO) strategy.}
  \label{fig:e_rep}
\end{figure}
\begin{figure*}[t]
  \centering
  \includegraphics[width=.96\linewidth]{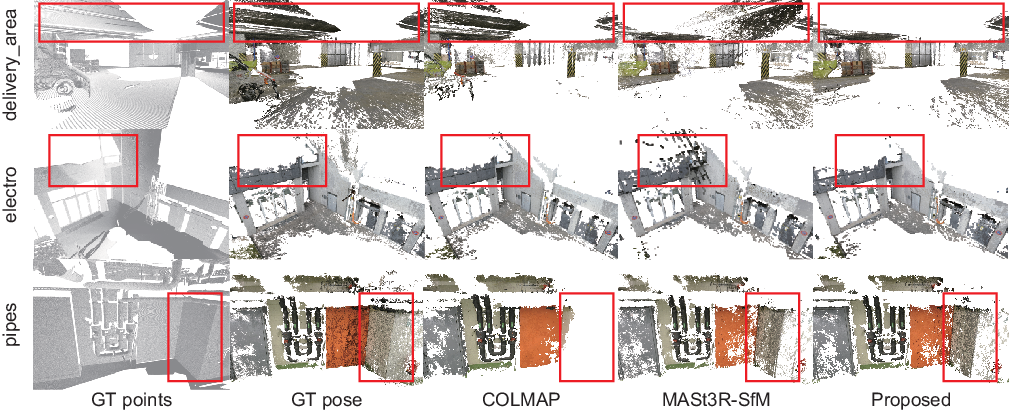}
  \caption{Qualitative comparison of 3D reconstruction results.
  The red boxes highlight that the proposed method effectively eliminates artifacts observed in MASt3R-SfM (top and middle rows) while successfully reconstructing texture-less regions where COLMAP fails (bottom row).}
  \label{fig:result_mvs}
\end{figure*}

\subsection{Camera Pose Estimation}
\label{exp:camera_pose_estimation}

Table \ref{tab:results_SfM} shows the quantitative results of camera pose estimation for each method.
We employ RRE, RTE, AUC@5, and SfM rate as evaluation metrics.
The reported values represent the average across the 13 scenes from the ETH3D dataset used in the experiments.
While MASt3R-SfM \cite{bardienus2024mast3r-sfm} achieves a 100\% SfM rate due to the robust matching capabilities of MASt3R \cite{vincent2024mast3r}, it exhibits relatively large average errors in RRE and RTE.
We attribute this to the inclusion of unreliable correspondences in the optimization process.
Both VGGSfM \cite{VGGSfM} and VGGT \cite{VGGT} tend to show large estimation errors.
Although VGGSfM achieves a higher AUC@5 than VGGT, its large average errors suggest the presence of extreme outliers in certain views.
In contrast, VGGT exhibits consistently large errors across the board.
On the other hand, our proposed method achieves the highest estimation accuracy among all methods, although the SfM rate decreases to 97\% due to the exclusion of unreliable views by the GVP module.
These results demonstrate that our method successfully optimizes camera parameters with high accuracy using the remaining views while effectively filtering out inappropriate ones.

Fig. \ref{fig:e_rep} shows the behavior of the reprojection error during the optimization process.
Our method converges to a lower final reprojection error compared to MASt3R-SfM.
Observing the error trajectory, although the error temporarily increases during the $Local$ and $Neighbor$ stages due to the addition of views, it significantly decreases in the subsequent $Global$ stage, converging to a lower value.
This result indicates that the iterative three-stage optimization prevents the solution from falling into local minima and refines parameters more effectively than a single optimization strategy using all views simultaneously.

\begin{table}[t]
  \centering
  \small
  \caption{Quantitative results of camera pose estimation on the ETH3D dataset.
  The best results are highlighted in bold.}
  \label{tab:results_SfM}
  \setlength{\tabcolsep}{3.5pt}
  \begin{tabular}{lcccc}
    \toprule
           & RRE $\downarrow$ & RTE $\downarrow$ & AUC@5 $\uparrow$ & SfM rate \\
    Method & [deg.]           & [deg.]           &  [\%]                & [\%]     \\
    \midrule
    COLMAP \cite{johannes2016colmap}          & 0.655            & 2.645            & 90.7             & 87       \\
    DFSfM \cite{DFSfM}                        & 2.298            & 3.711            & 68.0             & 85       \\
    VGGSfM \cite{VGGSfM}                      & 22.439           & 17.220           & 52.7             & 98       \\
    VGGT \cite{VGGT}                          & 2.485            & 8.806            & 35.4             & 100      \\
    MASt3R-SfM \cite{bardienus2024mast3r-sfm} & 2.572            & 3.343            & 75.7             & 100      \\
    \midrule
    G-MASt3R-SfM (Ours)                       & \textbf{0.474}   & \textbf{0.978}   & \textbf{93.9}    & 97       \\
    \bottomrule
  \end{tabular}
\end{table}

\subsection{Multi-View Stereo}

In this experiment, we evaluate the impact of the estimated camera parameters on 3D reconstruction.
We reconstruct 3D point clouds using MVSFormer++ \cite{MVSFormer++} with camera parameters obtained from COLMAP, MASt3R-SfM, and our proposed method (G-MASt3R-SfM).
Regarding the settings for MVSFormer++, we set the number of input images for depth map estimation to 10 and the depth confidence threshold to 0.5.
Additionally, we employ DPCD \cite{DPCD} for point cloud denoising.
To ensure a fair comparison, we resize input images to $640 \times 1,024$ pixels for all methods and scale the camera focal lengths accordingly.
Furthermore, only the views for which camera parameters were successfully estimated in Sect. \ref{exp:camera_pose_estimation} are used for 3D reconstruction.
Reconstruction accuracy is evaluated using the official ETH3D evaluation script after aligning the generated point clouds to the ground truth using the Iterative Closest Point (ICP) algorithm.

Table \ref{tab:results_MVS} presents the quantitative results.
Although COLMAP achieves the highest ``Acc.'', its low SfM rate limits the number of views available for reconstruction, resulting in lower ``Cpl.''.
In contrast, our proposed method records the highest ``Cpl.'' and also outperforms MASt3R-SfM in terms of ``Acc.''.
Consequently, our method achieves the best performance in ``F1 score'', which serves as the comprehensive metric.
This result indicates that our method accurately estimates camera parameters for each view while maintaining a sufficient number of views.

Fig. \ref{fig:result_mvs} shows examples of 3D reconstruction results by each method.
In ``{\tt delivery\_area}'' and ``{\tt electro}'', MASt3R-SfM generates noise in regions where no objects exist, due to incorrect camera pose estimation.
In contrast, since our method appropriately removes views with large errors by GVP, such noise is not observed, and the object shapes are correctly reconstructed.
Furthermore, in texture-less scenes like ``{\tt pipes}'', while COLMAP yields limited reconstruction coverage due to feature matching failures, MASt3R-based methods successfully reconstruct a wide range due to their dense matching capabilities.
These results demonstrate that our proposed method achieves both high-accuracy camera parameter estimation and robustness to texture-less regions, thereby improving the quality of 3D reconstruction.

\begin{table}[t]
  \centering
  \caption{Quantitative results of MVS on the ETH3D dataset.
  The best results are highlighted in bold.}
  \label{tab:results_MVS}
  \begin{tabular}{lccc}
    \toprule
    Method                                     & Acc. $\uparrow$ & Cpl. $\uparrow$ & F1 score $\uparrow$ \\
    \midrule
    COLMAP \cite{johannes2016colmap}           & \textbf{0.997}  & 0.631           & 0.729               \\
    MASt3R-SfM \cite{bardienus2024mast3r-sfm}  & 0.944           & 0.645           & 0.741               \\
    \midrule
    G-MASt3R-SfM (Ours)                        & 0.973           & \textbf{0.666}  & \textbf{0.762}      \\
    \bottomrule
  \end{tabular}
\end{table}

\section{Conclusion}
\label{sec:conclusion}

In this paper, we proposed G-MASt3R-SfM, a novel approach leveraging scene graphs to enhance the accuracy and robustness of MASt3R-based SfM.
By integrating the GVP and MSO modules, our method effectively filters geometrically inconsistent views and optimizes camera parameters through community-aware stepwise bundle adjustment.
Experiments on the ETH3D dataset demonstrated that our method achieves significantly improved performance in both camera pose estimation and 3D reconstruction quality compared to COLMAP and existing deep learning baselines.
In future work, we  will focus on refining the view selection algorithm and extending this framework to other parameter estimation tasks.

\section{Acknowledgment}
This work was supported in part by JSPS KAKENHI 23H00463 and 25K03131.

{\small
  \bibliographystyle{IEEEbib}
  \bibliography{paperlist}
}
\end{document}